\documentclass[conference]{IEEEtran}
\IEEEoverridecommandlockouts
\usepackage{cite}
\usepackage{amsmath,amssymb,amsfonts}
\usepackage{algorithm}
\usepackage{algorithmic}
\usepackage{subfigure}
\usepackage{caption}
\usepackage{graphicx}
\usepackage{bm}
\usepackage{textcomp}
\usepackage{xcolor}
\def\BibTeX{{\rm B\kern-.05em{\sc i\kern-.025em b}\kern-.08em
    T\kern-.1667em\lower.7ex\hbox{E}\kern-.125emX}}
\begin{document}

\title{Digital Twin Vehicular Edge Computing Network: Task Offloading and Resource Allocation\\
\thanks{This work was supported in part by the National Natural Science Foundation of China under Grant No. 61701197, in part by the National Key Research and Development Program of China under Grant No.2021YFA1000500(4), in part by the 111 Project under Grant No. B12018.}
}

\author{\IEEEauthorblockN{1\textsuperscript{st} Yu Xie}
\IEEEauthorblockA{\textit{School of Internet of Things Engineering} \\
\textit{Jiangnan University}\\
Wuxi, China \\
yuxie@stu.jiangnan.edu.cn}
~\\
\and
\IEEEauthorblockN{2\textsuperscript{nd} Qiong Wu*}
\IEEEauthorblockA{\textit{School of Internet of Things Engineering} \\
\textit{Jiangnan University}\\
Wuxi, China \\
qiongwu@jiangnan.edu.cn}
Corresponding Author
~\\
\and
\IEEEauthorblockN{3\textsuperscript{rd} Pingyi Fan}
\IEEEauthorblockA{\textit{Department of Electronic Engineering} \\
\textit{Tsinghua University}\\
Beijing, China \\
fpy@tsinghua.edu.cn}
}

\maketitle

\begin{abstract}
With the increasing demand for multiple applications on internet of vehicles. It requires vehicles to carry out multiple computing tasks in real time. However, due to the insufficient computing capability of vehicles themselves, offloading tasks to vehicular edge computing (VEC) servers and allocating computing resources to tasks becomes a challenge. In this paper, a multi task digital twin (DT) VEC network is established. By using DT to develop offloading strategies and resource allocation strategies for multiple tasks of each vehicle in a single slot, an optimization problem is constructed. To solve it, we propose a multi-agent reinforcement learning method on the task offloading and resource allocation. Numerous experiments demonstrate that our method is effective compared to other benchmark algorithms.
\end{abstract}


\section{Introduction}
With the development of Internet of Things, more and more in car applications are being installed in modern cars[1]. However, due to the insufficient computing capability of the vehicle itself, it may not be possible to handle the computing tasks generated by these vehicular applications in a timely manner. Vehicular edge computing (VEC), as a promising technology, can offload computing tasks to the VEC server or roadside unit, and assist in processing computing tasks by using the computing resources of the VEC server itself[2][3]. However, in such a way, VEC also faces new issues such as vehicle mobility and environmental dynamics. 

Digital twin, as an emerging technology, can bridge the gap between entities in physical space and digital space of information mapping[4],[5],[6]. Based on historical data and real-time status, establishing digital replicas for physical entities can enhance the real-time interaction, realize close monitoring, by reliable communication between digital space and physical systems[7].

In terms of vehicle task processing, applying digital twin technology to VEC task processing scenarios can better monitor vehicle operation status to control and or change it.In [8], Dai \textit{et al}. proposed an adaptive digital twin VEC network that utilizes digital twin caching content and designs an offloading scheme based on the DRL framework to minimize offloading latency. In [1], Cao \textit{et al}. applied digital twins to the task offloading scenario of VEC and solved the vehicle mobility problem by constructing a multiple objective joint optimization problem. In [9], Hevesli \textit{et al}. utilized DT to predict the real-time status of IoT devices in edge air-ground network for more efficient resource allocation. In [10], Tan \textit{et al}. applied modular processing of digital twins to the Internet of Vehicles, improving task scheduling efficiency and reducing energy consumption. However, they only considered the situation where the vehicle only generates one task in a single time slot. In practice, vehicles may have multiple demands at the same time, so it is possible to have multiple tasks in one time slot.

The main contribution of this paper is trying to simultaneously address the multiple service tasks and to construct a vehicle edge computing network with digital twin auxiliary task unloading processing, and propose an adaptive task unloading and resource allocation scheme\footnote{The source code has been released at: https://github.com/qiongwu86/Digital-Twin-Vehicular-Edge-Computing-Network\underline{ }Task-Offloading-and-Resource-Allocation} based on multi-agent reinforcement learning. The remaining parts of this article are as follows, Section \uppercase\expandafter{\romannumeral2} establishes and analyzes the system model, Section \uppercase\expandafter{\romannumeral3} constructs optimization problems and proposes solutions based on MDRL, Section \uppercase\expandafter{\romannumeral4} analyzes experimental results, and Section \uppercase\expandafter{\romannumeral5} summarizes.

\begin{figure}[!t]
	\centering
	\includegraphics[width=2.2in]{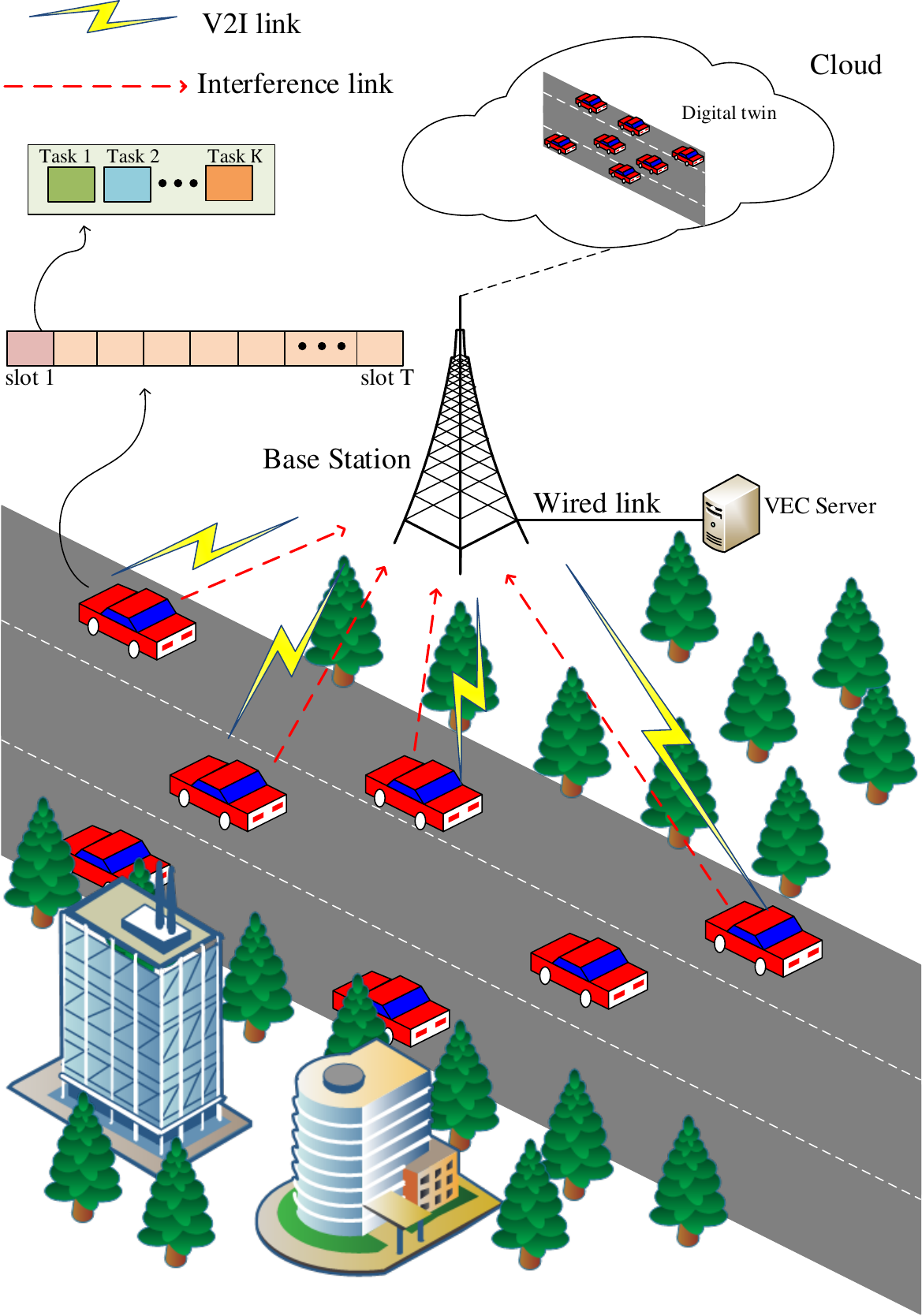}
	\caption{Digital twin vehicular edge computing network.}
	\label{figure1}
\end{figure}

\section{System model}
To simultaneously address multiple service tasks in VEC, in this section, we first describe the digital twin assisted multitask offload processing vehicular edge computing network model, and introduce the sub models.

As shown in Figure 1, we consider a DT scenario of VEC network in which a vehicle generates multiple tasks in a single time slot. It consists of two layers, namely the physical layer and the cloud digital twin layer. The physical layer includes $\mathbf{N}$ vehicles ($\mathbf{N}\stackrel{\Delta}{=} \{1,2,3,..., N\}$) and a single base station equipped with edge computing servers. The cloud digital twin layer is the digital mapping of each vehicle on the cloud. Specifically, each vehicle will generate $K$ tasks at a specific time slot. Due to the limitations of the vehicle's own computing resources, the digital twin in the cloud will make offloading decisions for the generated $K$ tasks to determine whether they are to be processed locally or offloaded. If they are to be offloaded to the VEC server for processing, the digital twin will also allocate computing resources for the processing of each task. It should be noted that we assume that the next task will only be generated after all K tasks for a single vehicle have been completed.

\subsection{Communication Model}
Due to our consideration of deploying digital twins in the cloud, we overlook the communication time between digital twins and base stations. In this model, we consider applying orthogonal frequency division multiple access (OFDMA) to $N$ V2I links, in other words, we have $N$ V2I links pre-allocated to the orthogonal frequency sub-bands, where $N$ V2I links occupy $N$ sub bands. Therefore, the communication rate $r_n ^t$between vehicle $n$ and base stations can be represented as
\begin{equation}
	r_n ^t = B\log_2 (1+\frac{p_n g_{n,b}}{\sigma^2}),
\end{equation}
where $B$ is each sub-band channel bandwidth, $p_n$ is the transmission power of vehicle $n$, $g_ {n, b} ^ t$ is the channel gain between vehicle $n$ and BS at time slot $t$, and $\sigma ^ 2$ is the additive noise power. We calculate the channel gain as follows
\begin{equation}
	g_{n,b}^t = |s_{n,b}^t|^2 h_{n,b}^t,
\end{equation}
where $s_{n, b} ^ t$ is small-scale fading, $h_{n, b} ^ t$ is large-scale path fading and shadows, where path fading is calculated as $128.1+37.6 \log_10(dis (v_n, bs))$, and $dis(v_n, bs)$ is the distance between vehicle $n$ and BS. For small-scale fading $s _ {n, b} ^ t$, we adopt a general fading model, as[11]
\begin{equation}
	s_{n,b}^t = \kappa s_{n,b}^{t-1} + \frac{1}{(\sqrt{(v_n^{pos-t}-bs^{pos})^2})^\alpha}
\end{equation}
where $s_{n, b} ^ t$ follows a circularly symmetric complex Gaussian distribution with unit variance, where $\kappa$ is a correlation coefficient, $v_n ^ {pos-t}$ is the position of vehicle $n$ in time slot $t$, $bs ^ {pos}$ is the position of the base station, and $\alpha$ is a path loss exponent.

\subsection{Computing Model}
Let $D_n=\{D_{n, 1}, D_{n, 2},... D_{n, k},... D_{n, K}\}$ represents the set of tasks generated by vehicle $n$, and for each task, we represent it as $D_{n, k}=\{d_{n, k}, c_ {n, k}, T_{n, k}\}$, where $d_{n, k}$ is the size of the $k$-th task generated by vehicle $n$, $c_{n, k}$ is the CPU cycle required to process a unit task, and $T_{n, k}$ is the maximum time limit required for the task. For the processing of the computing task $k$ of vehicle $n$, digital twin will give an offloading ratio of $\omega _ {n, k}$. When $\omega_{n, k} = 0$, it indicates local processing of the task, when $\omega _ {n, k} \in (0,1)$, it indicates partial offloading of the task, and when $\omega_{n, k} = 1$, it indicates complete offloading of the task. When vehicle $n$ generates tasks, if the $k$-th task $D_ {n, k}$ of vehicle $n$ needs to be processed locally, its local processing delay $t_{n, k} ^ {local}$ can be calculated as
\begin{equation}
	t_{n,k}^{local} = \frac{(1-\omega_{n, k})d_{n,k}c_{n,k}}{F_n},
\end{equation}
where $F_n$ is the computing resource of vehicle $n$ itself.
If there is an offloading operation in the $k$-th task of vehicle $n$, $D_{n,k}$, then the offloading part needs to be first transmitted to the base station equipped with the VEC server through wireless communication. Digital twin in the cloud estimates and allocates the required computing resources for the task, and returns the computing results to the vehicle. Here, due to the much smaller size of computing results compared to the upload part and the higher communication rate in the downlink, we do not consider the delay caused by the computing output results returning to the vehicle. In addition, it should be noted that due to the mapping relationship, there may be a positive and negative estimation error $\Delta f_ {n, k} ^ {est}$ between DT estimated computing resources and the actual situation[12],[13], resulting in a bias of the estimated time $\Delta t_{n, k} ^ {est}$, which can be defined as follows
\begin{equation}
	\Delta t_{n, k} ^ {est} = -\frac{\omega_{n, k}d_{n,k}c_{n,k}\Delta f_ {n, k} ^ {est}}{\widetilde{f}_{n,k}^{est}(\Delta f_ {n, k}+\widetilde{f}_{n,k}^{est})},
\end{equation}
Therefore, the calculation of the delay $t_{n, k} ^ {edge}$ generated by the edge execution of task $D_{n, k}$ can be given by
\begin{equation}
	t_{n, k} ^ {edge} = \frac{\omega_{n, k}d_{n,k}}{r_n^t} + \frac{\omega_{n, k}d_{n,k}c_{n,k}}{\widetilde{f}_{n,k}^{est}} + \Delta t_{n, k} ^ {est},
\end{equation}
where $\widetilde{f}_{n,k}^{est}$ represents the estimated computing resources for the DT of vehicle $n$ for task $D_{n, k}$.For task $D_{n, k}$, if there are partial offloading processing and local processing, it can be considered that both are executed in parallel. Therefore, the required delay $t_{n, k} ^ {exe}$ can be represented as
\begin{equation}
	t_{n,k}^{exe} = \max \{t_{n,k}^{local}, t_{n,k}^{edge}\},
\end{equation}
For a single vehicle $n$, the $K$ tasks generated are processed simultaneously, therefore, the total delay $t_n ^{total} $required to complete all executions can be calculated as
\begin{equation}
	t_n^{total} = \mathop{\max}\limits_{k\in K} (t_{n,k}^{exe}),
\end{equation}

\section{Problem formulation and solution}
\subsection{Problem Formulation}
The goal of this article is to minimize the total delay $t_n ^ {total}$ of $K$ tasks generated by each vehicle. After considering constraints such as VEC server computing resources and time constraints for each task, we formulate the problem as
\begin{subequations}\label{P1}
	\begin{equation}
		P1: \min t_n^{total}
	\end{equation}
	\begin{equation}
		\quad\qquad\qquad\qquad\Sigma_n ^N \Sigma_k ^K \Delta f_{n,k}^{est} +\widetilde{f}_{n,k}^{est} \le F_{max} ^{BS}
	\end{equation}
	\begin{equation}
		\quad\qquad\quad t_{n,k}^{exe} \le T_{n,k}, \forall n, k \in N, K
	\end{equation}
	\begin{equation}
		p_n \le p_n^{max}\quad
	\end{equation}
	\begin{equation}
		\omega_{n, k} \in [0,1] \qquad
	\end{equation}
\end{subequations}
Constraint (9b) means that the total computing resources required for task processing in the system cannot exceed the maximum computing resource $F_{max} ^ {BS}$ of the VEC server. Constraint (9c) means that the execution delay of task $D_{n, k}$ cannot exceed its maximum delay limit. Constraint (9d) means that the transmission power of vehicle $n$ cannot exceed its maximum transmission power $p_n ^ {max}$. Constraint (9e) indicates the offloading method of the task.

Due to the mutual influence of resource allocation among tasks in problem P1, which cannot be solved using conventional methods in polynomial time. Therefore, we use multi-agent methods to transform the problem. For multi-agent MDP, we use a tuple $(\mathcal{N}, \mathcal{S}, \mathcal{A}, \mathcal{R})$ to represent it, where $\mathcal{N}$ represents the set of agents, $\mathcal{S}$ represents the state space, $\mathcal{A}$ represents the action space, and $\mathcal{R}$ represents the reward function.
\subsubsection{\textbf{Agent Set} $\mathcal{N}$}
Let each vehicle act as an intelligent agent $n$, therefore, $N$ vehicles in the physical layer form the set of intelligent agents $\mathcal{N}={1,2,..., n,... N}$.
\subsubsection{\textbf{State Space} $\mathcal{S}$}
At each decision moment $t$, the state $s_n (t)$ that the intelligent agent $n$ can observe includes a set of $K$ tasks $D_n$ generated by the vehicle $n$, an estimation error $\Delta f _ {n, k} ^ {est}$, the position of vehicle $n$, and the channel gain between vehicle $n$ and BS. Therefore, the state of agent $n$ is represented as
\begin{equation}
	s_n (t) = \{ D_n, \Delta f_{n,k}^{est}, v_n ^{pos-t}, g_{n,b} ^t \},
\end{equation}
\subsubsection{\textbf{Action Space} $\mathcal{A}$}
After observing the current state $s_n (t)$, agent $n$ will take action $a_n (t)$. This action consists of the offloading ratio $\omega_{n, k}$ and the estimated computing resources $\widetilde{f}_{n,k}^{est}$ by DT. Therefore, the action of agent $n$ is represented as
\begin{equation}
	a_n (t) = \{\omega_{n, k}, \widetilde{f}_{n,k}^{est}\},
\end{equation}
\subsubsection{\textbf{Reward} $\mathcal{R}$}
Based on the current observed state $s_n (t)$, agent $n$ will take action $a_n (t)$ to interact with the environment, thereby obtaining a reward $r_n (t)$. Considering the constraints of the problem, we will express the reward as
\begin{equation}
	r_n (t) = \Sigma_{k=1} ^K \beta\frac{T_{n,k} - t_{n,k}^{exe}}{K} - \eta\frac{\Delta f_{n,k}^{est} +\widetilde{f}_{n,k}^{est} - F_{max}}{N},
\end{equation}
Where $\beta$ and $\eta$ are correlation coefficients used to control the two terms on the right side of the equation to the same order of magnitude. On this basis, we can express the long-term cumulative reward of agent $n$ as
\begin{equation}
	R_n(t) = \Sigma_{t_0 = 0} ^ t \gamma r_n(t_0),
\end{equation}
where $t_0$ represents the previous moment, and $\gamma\in(0,1) $represents the discount factor. By maximizing the long-term cumulative rewards of each agent, each agent $n$ can obtain the optimal execution strategy and resource allocation plan.

\begin{algorithm}[!t]
	\caption{Training Stage of MARL algorithm.}\label{alg:alg1}
	\renewcommand{\algorithmicrequire}{\textbf{Input:}}
	\renewcommand{\algorithmicensure}{\textbf{Output:}}
	\begin{algorithmic}[1]
		\REQUIRE{$D_n, \Delta f_{n,k}^{est}, v_n ^{pos-t}, g_{n,b} ^t$ for $n=1,...,N$ and $k=1,..., K$;}
		\ENSURE{$\theta_{\pi_n} ^{'}$ for $n=1,...,N$}
		\STATE Initialize discount factor $\gamma$ and parameter update rate $\eta$; 
		\STATE Random initialize the $\theta_{Q_n}, \theta_{Q_n} ^{'}, \theta_{\pi_n}, \theta_{\pi_n}^{'}$ for $n=1,...,N$;
		\FOR{$k$ eposide from 1 to $K$}
		\FOR{$n$ agent from 1 to $N$}
		\STATE Initialize $s_n (t)$;
		\STATE Input $s_n (t)$ to estimation actor network, and get $a_n (t)= \pi_n (s_n(t);\theta_{\pi_n})$;
		\STATE Execute $a_n(t)$ based on $s_n(t)$, obtain $r_n(t)$ and transfer to $s_n(t+1)$;
		\STATE Store $(s_n(t), a_n(t), r_n(t), s_n(t+1))$ as an experience in the Replay Buffer;
		\STATE Input $\mathcal{S}$ and $\mathcal{A}$ to estimation critic network and compute $Q_n(\mathcal{S},\mathcal{A};\theta_{Q_n})$;
		\STATE Input $\mathcal{S^{'}}$ ,$\mathcal{A^{'}}$ to target critic network and compute $Q_n ^{'}(\mathcal{S^{'}},\mathcal{A^{'}};\theta_{Q_n}^{'})$;
		\STATE Calculate Q value by(14),temporal difference $\delta$ and loss function $L(\theta_{Q_n})$ by (16);
		\STATE Update $\theta_{Q_n}$ by stochastic gradient descent,i.e. (17);
		\STATE Input $s_n(t)$ to estimation actor network and obtain $a_n(t)=\pi_n (s_n(t);\theta_{\pi_n})$;
		\STATE Input $s_n(t+1)$ to target actor network and obtain $a_n(t+1)=\pi_n^{'}(s_n(t+1);\theta_{\pi_n}^{'})$;
		\STATE Update $\theta_{\pi_n}$ by gradient descent, i.e. (18)
		\STATE Update $\theta_{Q_n}^{'}$ and $\theta_{\pi_n}^{'}$ by (19) and (20), respectively;
		\ENDFOR
		\ENDFOR
	\end{algorithmic}
	\label{alg1}
\end{algorithm}

\subsection{Solution}
We propose a MADRL algorithm to learn how to maximize long-term cumulative rewards. The algorithm structure includes an actor network and a critic network, where the actor network consists of an estimation actor network and a target actor network. Similarly, a critic network consists of an estimation critic network and a target critic network. The algorithm is divided into centralized training and distributed execution. Specifically, since the server manages the critical network of each agent, it can obtain the status and actions of all agents. During the centralized training phase, the server first obtains the status and actions of all agents. Therefore, the server can train the estimated critical network for each agent, thereby achieving the goal of maximizing the Q-value. Distributed execution means that agent $n$ observes and obtains local states, and executes actions based on policy $\pi_n$. It should be noted that in the early stages, due to the lack of experience, agent $n$ will randomly take actions to explore more possible actions. When the experience is sufficient, agent $n$ will take actions to maximize rewards. 
For a single agent $n$, The ways to obtain and update the Q value of the critical network are
\begin{equation}
		Q_n (\mathcal{S},\mathcal{A};\theta_{Q_n})=\mathbb{E}_{\pi} [R_n (t);s(t),a(t);\theta_{Q_n}],
\end{equation}
and
\begin{equation}
	Q_n (\mathcal{S},\mathcal{A};\theta_{Q_n})=R_n (t)+\gamma_n \max Q_n ^{'}(\mathcal{S}^{'},\mathcal{A}^{'};\theta^{'} _{Q_n}),
\end{equation}
where $\theta_{Q_n}$ and $\theta^{'} _{Q_n}$ are the parameters for estimation critic network and target critic network, respectively. Eqn. (18) is derived from the Bellman equation[14].

The critical network is controlled by parameters $\theta_{Q_n}$, and to obtain the optimal parameters, it is necessary to know the loss function. Here, the loss function can be expressed as
\begin{equation}
	L(\theta_{Q_n})=\mathbb{E}(\delta ^2),
\end{equation}
where $\delta $ is the time difference error, calculated as $\delta = Q_n ^{'}(\mathcal{S}^{'},\mathcal{A}^{'};\theta^{'} _{Q_n}) - Q_n (\mathcal{S},\mathcal{A};\theta_{Q_n})$.
To minimize the loss function, we update the critic network parameters $\theta_{Q_n}$ through random gradient descent
\begin{equation}
	\nabla_{\theta_{Q_n}} L(\theta_{Q_n}) = \mathbb{E}(2\delta \nabla_{\theta_{Q_n}} Q_n(\mathcal{S},\mathcal{A};\theta_{Q_n})),
\end{equation}

Due to the distributed execution of the actor network, which is deployed on each vehicle, the agent $n$ takes action by observing local states. For the parameter update method of the actor network, we choose gradient descent
\begin{equation}
	\nabla_{\theta_{\pi_n}} L(\theta_{\pi_n}) \approx \mathbb{E}[\nabla_{\theta_{\pi_n}} \log \pi_n (s_n (t );\theta_{\pi_n}) Q_n (\mathcal{S},\mathcal{A};\theta_{Q_n})],
\end{equation}
where $\pi_n (s_n (t); \theta_ {\pi_n})$ represents the action strategy adopted by agent $n$ based on the current state.

During the training process, in order to ensure the stability of the algorithm, we adopt a soft update method for the parameters of the target actor network and target critic network
\begin{equation}
	\theta_{Q_n} ^{'} = \epsilon \theta_{Q_n} + (1-\epsilon)\theta_{Q_n} ^{'},
\end{equation}

\begin{equation}
	\theta_{\pi_n} ^{'} = \epsilon \theta_{\pi_n} + (1-\epsilon)\theta_{\pi_n}^{'},
\end{equation}
where $\epsilon$ is parameter update rate, which $\epsilon \in [0,1]$. $\theta_{\pi_n} ^{'}$ and $\theta_{\pi_n}$ are the parameters for estimation actor network and target actor network, respectively.

The specific process of the algorithm is described in Algorithm 1.

\section{Simulation results}
The simulation experiment in this article was implemented using Python and PyTorch. In terms of environmental settings, the settings of environmental parameters are shown in Table \uppercase\expandafter{\romannumeral1}. Each sub-band bandwidth $B$ is set to 50MHz, the transmission power of the vehicle $p_n$ is set to 200mW, the noise power $\sigma ^ 2$ is set to $10^{-11}mW$, the correlation coefficient $\kappa$ is 0.2 and the path attenuation exponent $\alpha$ is 2. The number $K$ of computing tasks generated by the vehicle in a single time slot is 3. The size $d_{n,k}$ of each computing task is randomly selected in [100, 150] bytes, and the required computing resource per unit size task $c_{n,k}$ is 0.25MHZ/byte. The computing time limit $T_{n,k}$ for the computing task is set to 0.2s. The computing resources of the server $F_{max} ^{BS}$ and the vehicle itself $F_n$ are 100GHz and 5GHZ respectively. The DT mapping error $\Delta f_{n,k}$ is randomly selected in [-0.5,0.5], and the balance factor $\eta$ and $\beta$ are set to 0.5 and 10, respectively. The discount factor $\gamma$ is set to 0.95.

\begin{table}[h]
	\caption{Experimental parameter settings\label{tab:table1}}
	\renewcommand\arraystretch{1.5}
	\centering
	\begin{tabular}{|c|c|c|c|}
		\hline
		\textbf{Parameter} & \textbf{Value} & \textbf{Parameter} & \textbf{Value}\\
		\hline
		$B$ & 50MHZ & $p_n$ & 200mW \\
		\hline
		$\sigma ^ 2$ & $10^{-11}mW$ & $\kappa$ & 0.2\\
		\hline
		$\alpha$ & 2 & $K$ & 3\\
		\hline 
		$d_{n,k}$ & [100, 150]bytes & $c_{n,k}$ & 0.25MHZ/byte\\
		\hline
		$T_{n,k}$ & 0.2s & $F_{max} ^{BS}$  & 100GHZ\\
		\hline
		$F_n$ & 5GHZ & $\Delta f_{n,k}$ & [-0.5,0.5]\\
		\hline
		$\eta$ & 0.5 & $\beta$& 10\\
		\hline
		$\gamma$ & 0.95 & $\epsilon$ & 0.01\\
		\hline
	\end{tabular}
\end{table}

%\subsection{Performance Evaluation}
In Figure 2, as the number of iterations increases, the MARL algorithm eventually converges to a larger value, but the SAC algorithm always fluctuates. This is because the SAC algorithm actually only updates actions based on local observations by one agent at each epoch $t$, while the actions of other agents remain unchanged. At the same time, a single DQN is shared among all agents, making it impossible to learn suitable schemes.

\begin{figure}[!t]
	\centering
	\includegraphics[width=2.3in]{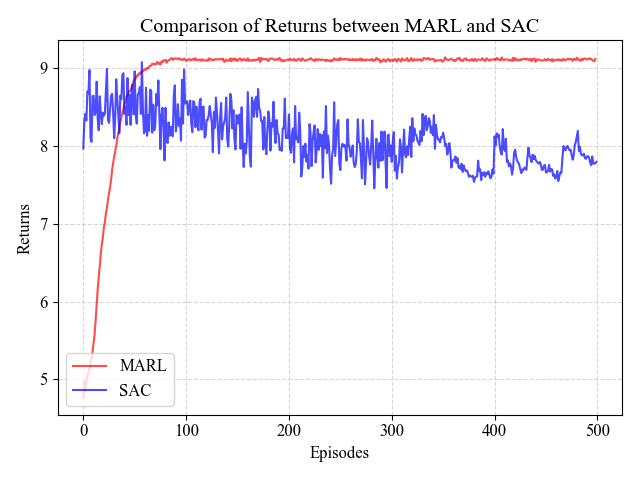}
	\caption{Comparison of average rewards between different algorithms:$N=4, p_n=200mW, \Delta f=0.2$.}
	\label{figure2}
\end{figure}

\begin{figure}[htbp]
	\centering
	\begin{subfigure}[resource allocation]{
			\centering
			\includegraphics[width=2.3in]{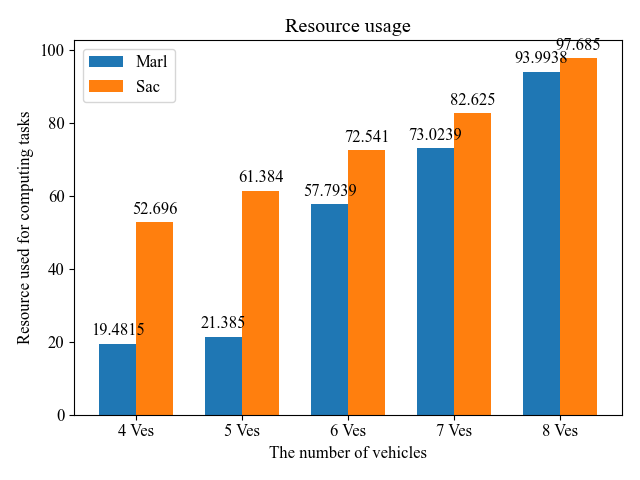}
			%\caption{dt}
			\label{figure3.a}}
	\end{subfigure}
	\centering
	\begin{subfigure}[Task processing latency and resource allocation of different algorithms.]
		{
			\centering
			\includegraphics[width=2.3in]{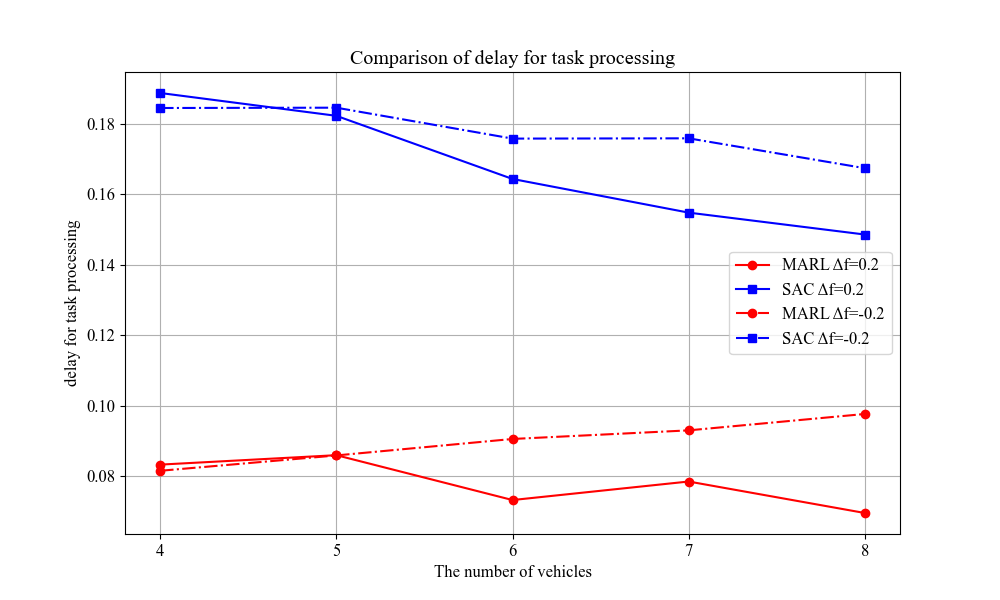}
			%\caption{tk}
			\label{figure3.b}}
	\end{subfigure}
	\caption{Comparison of task processing delays and resource utilization.}
	\label{figure3}
\end{figure}
In Figure 3(a), we compared the performance of the Marl and SAC algorithms in resource utilization. It can be seen that the Marl algorithm uses fewer computing resources compared to the SAC algorithm. This is because when facing multiple tasks for a single vehicle, the Marl algorithm can flexibly allocate computing resources according to the size of each task based on global information. In addition, as the number of vehicles increases, the number of computing resources used by both algorithms will increase. This is because the increase number of vehicles will also increase the number of tasks that need to be processed in a single time slot. In order to not exceed the task deadline, vehicles will use more resources. In Figure 3(b), we compared the task processing delay of two algorithms under different estimation errors. Specifically, mapping errors between digital twins and vehicles can cause some processing delays. When the error is positive, i.e $\Delta f = 0.2$, the digital twin overestimates the resource allocation required for task processing, thereby allocating more computing resources to the computing task, which naturally reduces the computation delay. On the contrary, when the error is negative, i.e $\Delta f =-0.2$ the resources obtained by the vehicle's computational task decrease due to the underestimation of digital twins, resulting in an increase in computational delay.

\section{Conclusion}
This article considered a VEC network scenario with digital twin assisted offloading and resource allocation, and constructed an optimization problem for vehicle task offloading and server resource allocation by analyzing the offloading process in this scenario. Due to the non-convexity of this problem, we used multi-agent MDP to rephrase the problem and proposed an algorithm based on MARL to obtain the optimal task offloading and resource allocation strategy. The conclusion is summarized as follows
\begin{itemize}
	\item Our method can obtain the much better offloading strategy for vehicles generating multiple tasks in a single time slot comparing some benchmark methods.
	\item  Our method utilizes digital twins to intelligently allocate real-time computing resources from VEC servers to multiple tasks generated by vehicles in a single time slot and multi task VEC scenario. It may provide more new insights to the VEC applications in the future.
\end{itemize}

\end{document}